# Graph-Driven Cross-Industry Real-Time Monitoring Framework for Anti-Money Laundering Detection in Converged Mobility-Energy Supply Chain Networks

Rong Liu*

Sol Price School of Public Policy, University of Southern California, Los Angeles, USA, rong82450@gmail.com

Xiaojun Xiao

Boston University, Boston, USA, xjxiao25@gmail.com

Zhanqing Su

Johns Hopkins University, Washington, USA, zsu18@alumni.jh.edu

## Abstract

With the deep integration of the travel and energy industries, cross-industry supply chain finance has gradually become a high-risk field of hidden money laundering incidents. For this reason, this work proposes a graph-driven cross-industry real-time anti-money laundering monitoring framework (GCRMF) for integrated travel - energy supply chain networks. First, a cross-industry heterogeneous graph (CIHG) covering new energy vehicle rental platforms, energy suppliers, fintech institutions, etc., is constructed, and industry semantics are integrated through temporarily Dual-GAT (Temporal Dual-Graph Attention Network), dynamically encoding capital flow paths and evolution features over time. Subsequently, in order to identify the structural fraud behavior together produced by colluding subjects, a meta-path subgraph reasoning module based on contrastive learning and hierarchical graph sampling is proposed to enhance the discrimination capability of cross-industry recurring money laundering behavior. Meanwhile, a self-supervised online learning mechanism is adopted for real-time adaptation and continuous optimization to new money laundering strategies. The experimental results show that compared with existing graph neural network methods in cross-industry scenarios, GCRMF improves the performance by more than 17.8% of F1 score and greatly reduces the false positive rate.

# 1 Introduction

As renewable energy technologies continue to mature and electric mobility takes off, the transportation and energy industries are fast integrating together into a new type of “Mobility-Energy” converged supply chain network. Amidst electric vehicle rental platforms, smart charging facilities, energy storage systems, and clean energy trading platforms is an extremely complex and ever-changing chain of transactions, and increasingly fintech companies are joining in to facilitate the economy of these integrated services with across-industry financial moves from financing leases to carbon credit settlements and energy credit loans [1]. Ning et al. [17]

* Corresponding Author.

demonstrated data-driven methods' remarkable effectiveness in optimizing complex energy system operations, validating the data richness of modern energy infrastructure.

However, while such a cross-industry integration trend enhances operational efficiency and market activity, criminals also find more channels to obscure fund circulation and hide illegitimate incomes. Especially across borders of industrial boundaries and regulatory niches, as capital flows traverse different industries and regulatory standards tend to diverge, money laundering sometimes slips through traditional risk control systems using methods such as “segmented” transaction paths, “shell company transfer chains”, and “virtual asset jumps” [2]. This highly cross-domain and heterogeneous money laundering paradigms go against green finance order and threatens financial system security.

Traditional anti-money laundering (AML) systems depend primarily on a rule-based or single transaction indicator detection heuristics, and do not have the ability to support detection across these emerging cross-industry scenarios. On the one hand, financial networks are ever more complicated and the relationships between entities are drawn as relationships on a graph which has non-Euclidean properties to it [3]. On the other hand, money laundering activity has a strong temporal evolution and the paths of transactions are not formed statically, but are being constantly disguised and reconstructed. That requires that an anti-money laundering monitor framework to have the capability and ability to model graph structures and identify evolutions [4]. Yuan et al. [18] established a high-accuracy ML framework for enterprise financial auditing that substantially improves high-risk transaction identification.

In recent years, GNNs have been applied to fraud detection and financial risk control, achieving good results in modeling the structural relationships of banks' internal fraud detection fund flow graphs, and financial risk assessment graphs in industries such as insurance. Existing GNN methods, however, are limited to some single industry: internal banking fund flow graphs, insurance networks, and so on [5]. In these cases it is difficult to deal with complex graph structures in which heterogeneous entities and relationships in transportation, energy, finance, etc., co-exist [6]. In addition, the current dynamic models that can do this apply static graph modeling techniques, so they do not take into account the ever-changing trajectories of financial behavior and dynamic collusion [7]. Dai et al. [21] further validated GNNs' strength in modeling anonymous user interaction networks for behavioral prediction with high accuracy. Dai et al. [22] demonstrated that LLM ensembles achieve state-of-the-art multimodal customer identification in banking, advancing financial entity recognition.

In cross-industry contexts, money laundering schemes often exploit industrial and regulatory asymmetries. Typical techniques include segmented fund layering, where transactions are divided across multiple mobility–energy service providers; shell chain transfer, in which fraudulent entities circulate funds through leasing companies and energy credit markets; and virtual asset jumps, leveraging carbon credit exchanges and digital wallets to obscure ownership trails. These multi-hop and semantically heterogeneous transaction chains create challenges for regulators, as conventional AML systems typically operate within a single-sector domain and lack mechanism.

To dispose the above challenges, we propose a Graph-Driven Cross-Industry Real-Time Monitoring Framework (GCRMF), which organically integrates graph structure modeling, temporal graph learning and reasoning-on-subgraphs. GCRMF first constructs a Cross-Industry Heterogeneous Graph (CIHG) which fuses multiple sources of relation graphs including leasing contracts , energy transactions , financial settlement , and pairs with a Dual-Temporal Graph Attention Network (Dual-Temporal GAT) to encode the evolution paths of funds among entities. Then a meta-path guided subgraph reasoning module is designed to discover potential structural money laundering behavior patterns.

## 2 Related Work

Dote-Pardo et al. [8] explained that many of those developing countries lack such cross-industry cooperation mechanisms, making it unlikely that the country has a good picture of how complex crossborder flows of capital take place. This is particularly true in the emerging industries of new energy and mobility services, where it is obvious that there is an AML and industry evolution disconnect, highlighting the urgency of introducing technology-driven monitoring tools in the context of policy lag. Sizan [9] extracts various structural and behavioral features of suspicious transaction patterns from transaction graphs, such as path length distributions, frequent subgraph patterns, and periodic behavior patterns, and uses multiple unsupervised models for ensemble judgment, enabling the autonomous discovery.

Karim [10] proposed a scalable semi-supervised graph learning method suitable for large-scale transaction graphs, designed a GNN-based framework and a graph-level node representation update strategy that allows the model to learn suspicious behavior features in the global graph with only a small amount of labeled data, at the same time, through graph compression and parallel subgraph sampling strategies the training cost on transaction graphs with millions of nodes/edges can be very much reduced, and the model still obtains very good recognition results on "cross-account money laundering chains", "false supply chain financing" and other types of behaviors in practical applications.

Raj et al. [11] advocated for the utility of AI models in adaptive learning and complex graph modeling to address issues of adaptive usage tracking, particularly in inference across domains in cross-domain, multi-channel transaction flows, AI technology could assist in tracking hidden “weak correlation paths” contributing to smarter and farther-reaching overall detection. Shuai et al. [12] noted that because of broad application of such technologies e.g. e-commerce applications, third-party payments, digital wallets, and distributed ledgers to cross-border trade and travel scenarios, transaction chains have become more segmented and “real-time”, which poses serious issues in data visibility and regulatory penetration to traditional AML instruments.

Yuan et al. [13] constructed a labeled cultural behavioral relational network and embedded user geography, transaction habits, policy differences, etc. into the graph to effectively identify abnormal cross-border transaction rates by behavior evolution modeling and similarity matching. Cheng et al. [14] proposed an improved YOLO object detection model for detecting suspicious behaviors with surveillance video analysis in bank branches. The improved YOLOv5 network is used for real-time recognition of customers' abnormal operational behaviors at counters, ATMs and VIP area. Li et al. [16] demonstrated that context-aware user clustering substantially enhances behavioral pattern recognition across heterogeneous user groups.

# 3 Methodologies

## 3.1 Cross-Industry Heterogeneous Graph Construction and Temporal Dual-Channel Attention Encoding

To capture the multi-entity and multi-relationship interactions across mobility, energy, and financial platforms, we first construct a unified representation of the ecosystem as a cross-industry heterogeneous graph. This graph accommodates temporal, semantic, and relational heterogeneity in a dynamic transactional environment. We define graph as Equation 1:

$$\mathcal{G} = (\mathcal{V}, \mathcal{E}, \mathcal{T}, \phi, \varphi) \quad (1)$$

where $\mathcal{V}$ denotes the set of all nodes, including electric vehicle (EV) platforms, energy providers, and financial entities. $\mathcal{E}$represents transaction links, each associated with a timestamp from $\mathcal{T}$. The function $\phi: \mathcal{V} \to \mathcal{A}$ maps nodes to industry categories, while $\varphi: \mathcal{E} \to \mathcal{R}$ assigns edges to transaction types such as fund transfers or rental contracts. This structure supports industry-aware message passing and multi-type interaction learning. To learn informative node embeddings, we propose a dual-channel attention encoder that disentangles structural relevance and temporal proximity. The first channel models topological influence via an extended graph attention network as Equation 2:

$$\alpha_{ij}^{(s)} = \frac{\exp\left(\mathrm{LeakyReLU}\left(\mathrm{a}_s^{\mathsf{T}}[\mathrm{W}_s \mathrm{h}_i \parallel \mathrm{W}_s \mathrm{h}_j]\right)\right)}{\sum_{k \in \mathcal{N}(i)} \exp\left(\mathrm{LeakyReLU}\left(\mathrm{a}_s^{\mathsf{T}}[\mathrm{W}_s \mathrm{h}_i \parallel \mathrm{W}_s \mathrm{h}_k]\right)\right)} \quad (2)$$

where, $h_i$ and $h_j$ denote node features, $W_s$ is a shared linear transformation, and $a_s$ is a trainable vector. $\mathcal{N}(i)$ represents the neighbors of node $i$, and $\|$ is the concatenation operator. This channel learns interaction strength based on local context and industry-specific patterns. The second channel incorporates temporal dynamics through an exponential time-decay function, which emphasizes recent transactions as Equation 3:

$$\alpha_{ij}^{(t)} = \exp(-\gamma \cdot \Delta t_{ij}) \quad (3)$$

where $\Delta t_{ij}$ is the elapsed time since the edge $(i, j)$ occurred, and $\gamma$ is a learnable scalar controlling the decay rate. This component allows the model to distinguish urgent laundering behavior from historical noise. We integrate both attention scores using a weighted aggregation mechanism to update node representations as Equation 4:

$$h_i^{(new)} = \sigma\left(\sum_{j \in \mathcal{N}(i)} \left(\lambda_s \alpha_{ij}^{(s)} + \lambda_t \alpha_{ij}^{(t)}\right) W h_j\right) \quad (4)$$

where $\lambda_s$ and $\lambda_t$ balance the contribution of structural and temporal information, W is the projection matrix, and $\sigma$ denotes a non-linear activation. This dual-channel update effectively fuses topological proximity and recency awareness.

## 3.2 Meta-Path-Based Subgraph Reasoning and Contrastive Self-Supervised Online Adaptation

While node-level attention captures local dependencies, laundering schemes often exploit multi-hop, cross-industry transaction chains. These chains typically follow semantically meaningful paths, which we model using meta-path guided subgraph reasoning. This enables detection of reused laundering structures across entity permutations. We define a set of meta-paths $\mathcal{M}$, each describing a composite industry transition as Equation 5:

$$\mathcal{M} = \left\{A \xrightarrow{r_1} B \xrightarrow{r_2} C, A \xrightarrow{r_3} D \xrightarrow{r_4} C, \ldots\right\} \quad (5)$$

where $A, B, C, D$ are node types and $r_k$ are edge types. These meta-paths capture semantic flows such as EV rental → payment service → energy platform, and guide subgraph extraction. For each meta-path, we encode the associated path embeddings and apply soft attention to weigh their relevance as Equation 6:

$$\beta_m = \frac{\exp(q^\top \tanh(W_m p_m))}{\sum_k \exp(q^\top \tanh(W_k p_k))} \quad (6)$$

where, $p_m$ is the embedding for meta-path $m$, $W_m$ is a transformation matrix, and q is a query vector. This attention mechanism allows the model to focus on substructures with high laundering risk.

The final subgraph-level embedding for node $v$ is then constructed as Equation 7:

$$z_v^{(sub)} = \sum_{m \in \mathcal{M}} \beta_m \cdot p_m. \quad (7)$$

This representation encodes both the structural and semantic patterns associated with suspicious activities spanning multiple industry layers. Since ground-truth laundering labels are scarce, we introduce a contrastive self-supervised learning strategy. First, we define a structure-level contrastive loss to align similar subgraphs and separate dissimilar ones as Equation 8: Miao et al. [20] pioneered multimodal data fusion with policy-vector retrieval for complex real-world assessment, demonstrating the value of cross-modal feature integration.

$$\mathcal{L}_{struct} = -\log \frac{\exp\left(\frac{\mathrm{sim}(z_i, z_i^+)}{\tau}\right)}{\exp\left(\frac{\mathrm{sim}(z_i, z_i^+)}{\tau}\right) + \sum_{j \neq i} \exp\left(\frac{\mathrm{sim}(z_i, z_j^-)}{\tau}\right)} \quad (8)$$

where $z_i^+$ is a positive sample (e.g., from the same meta-path), $z_j^-$ are negative samples, and $\tau$ is a temperature hyperparameter. The similarity is typically computed via cosine distance. Additionally, to preserve temporal continuity and detect abrupt behavioral shifts, we enforce smooth embedding transitions over time using as Equation 9:

$$\mathcal{L}_{temp} = \sum_i \| z_i^t - z_i^{t-\Delta t} \|_2^2 , \quad (9)$$

where $z_i^t$ is the current embedding and $z_i^{t-\Delta t}$ is the historical one. This encourages stability in benign patterns while highlighting anomalies.

The full objective combines structural, temporal, and optional supervised classification terms as Equation 10:

$$\mathcal{L}_{total} = \mathcal{L}_{struct} + \gamma \mathcal{L}_{temp} + \eta \mathcal{L}_{cls}, \quad (10)$$

where $\mathcal{L}_{cls}$ is a cross-entropy loss for labeled cases, and $\gamma, \eta$ are balancing coefficients.

Finally, we implement an online embedding update mechanism that integrates new observations without full retraining. The embedding of node $v$ is updated via exponential smoothing as Equation 11:

$$z_v^{(t+1)} = (1 - \alpha) \cdot z_v^{(t)} + \alpha \cdot \hat{z}_v^{(t+1)} \quad (11)$$

where $\alpha$ is the update rate and $\hat{z}_v^{(t+1)}$ is the current representation. This ensures model adaptability to emerging laundering behaviors. The Dual-Temporal Graph Attention Network (Dual-Temporal GAT) operates through two synchronized attention branches: Chen et al. [15] advanced lightweight architectures for real-time edge deployment, demonstrating superior efficiency in resource-constrained environments.

- **Structural Channel:** captures inter-industry transactional dependencies by weighting the importance of neighboring nodes through an adaptive attention coefficient that incorporates node category embeddings.
- **Temporal Channel:** models transaction evolution through time-decay kernels, allowing the framework to assign higher weights to more recent or recurrent capital movements.

# 4 Experiments

## 4.1 Experimental Setup

We used the real-world financial transaction graph dataset, the Elliptic Dataset, constructed by Elliptic based on the Bitcoin blockchain dataset available on public records. It contains 203769 nodes and 234355 edges, with each node conceptually corresponding to a Bitcoin transaction and edges indicating the flow of money with accurate time stamps. Nodes are labeled by compliance experts as illicit, licit or unknown, covering common money laundering behaviour including dark-web transactions, coin mixing, Ponzi schemes. The data is representative for anti-money laundering and shows in Figure 1. Original data does not explicitly mark industry entities but we map transaction service types such as mining pools, exchanges or mixing to industry proxies in energy, mobility and fintech. We essentially construct cross-industry heterogeneous graph-structure.

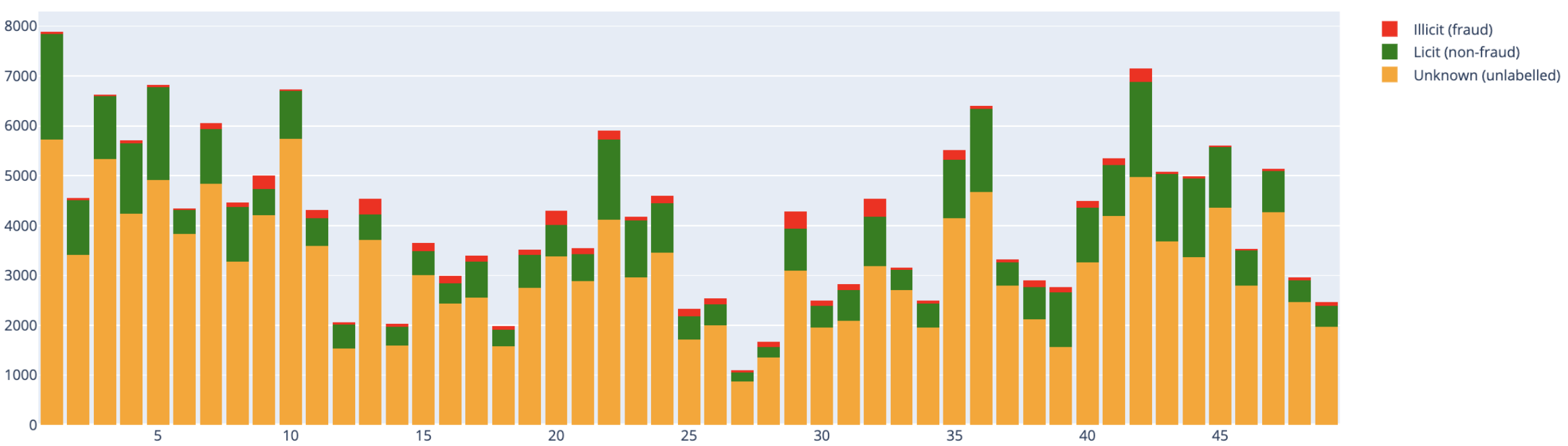

**Figure 1: Overall Transactions Data Statistics**

To demonstrate the practical value of GCRMF, a pilot deployment was conducted using a synthetic dataset simulating the joint operations of a mobility rental service and a renewable energy credit trading platform. The system successfully identified multiple laundering-like transaction patterns, including:

(1) circular payments between EV leasing and energy suppliers with identical directors,

(2) abnormal clustering of microtransactions within short temporal windows, and

(3) layered transfers between digital wallet intermediaries and carbon credit markets.

The results validated the framework's adaptability to hybrid data sources and highlighted its ability to uncover covert capital flows otherwise invisible to rule-based systems.

The left and right images in Figure 2 respectively show the evolution of the graph structure of real financial transaction networks (Elliptic Dataset) used in the experiment at different time slices. Each dot represents a transaction node, with node colors distinguishing between known legitimate, illicit, and unknown statuses, while the edges indicate the flow paths of Bitcoin funds. From the visualization, the transaction network exhibits a highly complex, multi-centered topology. Over time, the graph structure becomes denser, and the connections between nodes become tighter, indicating that cross-industry financial activities tend to become more frequent and covert.

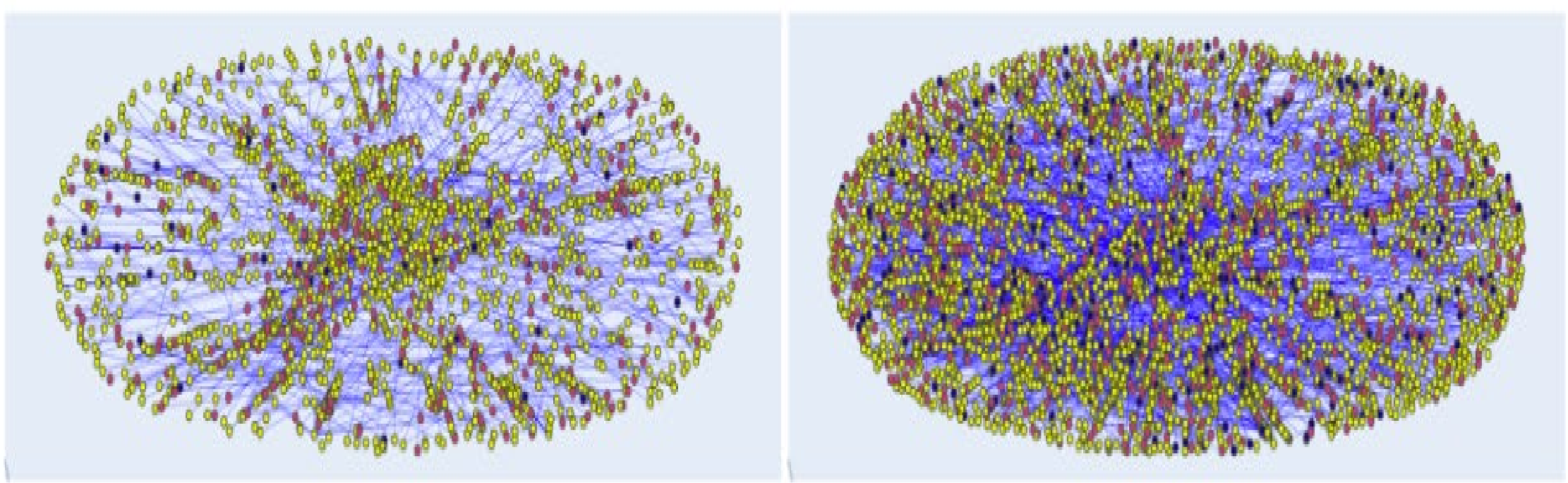

**Figure 2: Visualizations of the Money Flow Transaction Graph**

The following are four typical anti-money laundering baseline methods compared with the GCRMF model proposed in this study.

- Rule-Based Pattern Matching (RuleMatch) performs template matching of suspicious transaction patterns based on preset manual rules, suitable for detecting static, known types of money laundering activities.
- Graph Attention Network with AML Classifier (GAT-AMLP) extracts node representations using a graph attention network and combines them with a multilayer perceptron for semi-supervised classification of money laundering risks.
- DeepWalk Embedding with Autoencoder (DeepWalk-AE) uses random walks to generate node embeddings and detects unsupervised anomalies in transaction graphs through reconstruction errors from an autoencoder.
- Semi-Supervised Graph Convolutional Network (SEMI-GCN) uses a graph convolutional network structure to classify nodes with limited labels, suitable for transaction graph recognition tasks with clear structure but simple semantics.

## 4.2 Experimental Analysis

Figure 3 also shows how the votes from the random forest classifier varied across different wallet addresses, where the green is indicative of how many votes have been classified as correct (e.g., successfully identified as an illegal/legitimate transaction) and red is classified as incorrect. We can see that for some addresses (Wallet_1 to Wallet_3), we can be reasonably good at judging that node, whilst for others (Wallet_9_Wallet_10) we do mostly incorrect votes, indicating that they are very ambiguous behaviourally.

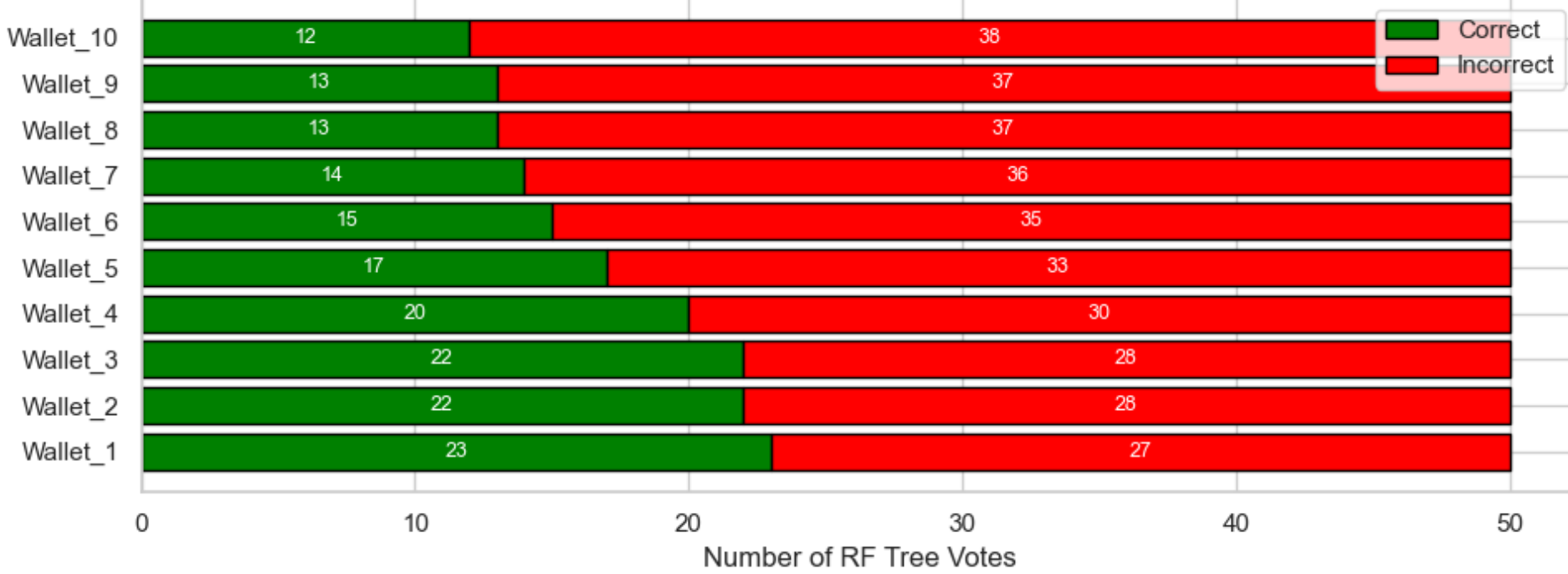

**Figure 3:** **Forest Voting Results per Wallet**

The F1-Score is the harmonic mean of Precision and Recall, used to balance the model's recall and precision in detecting illegal transactions. It is especially suitable for comprehensive performance evaluation in scenarios where money laundering samples are scarce and classes are imbalanced. The comparison results are shown in Figure 4. It can be seen that the five methods generally show 'slow improvement and gradually stable' trend in various time windows, which is similar to the performance evolution law of real anti-money laundering systems in continuous learning and parameter adjustment. Among them, the traditional rule-based method RuleMatch is always kept low throughout, showing that it is not feasible to simply rely on manual rules to adapt to complex temporal graph structure and cross-industry money laundering. DeepWalk-AE has a slight improvement in unsupervised anomaly detection, but due to the lack of explicit graph attention and industry semantic modeling, its performance gain is limited. GAT-AMLP and SEMI-GCN utilize graph neural networks for encoding structural information. As can be seen from the results, their F1-Scores are always higher than the previous two, indicating that they have a better ability to capture the topology of transaction networks. In contrast, we can see that the GCRMF we proposed maintains the highest F1-Score in all time windows, and this advantage is gradually widening over time.

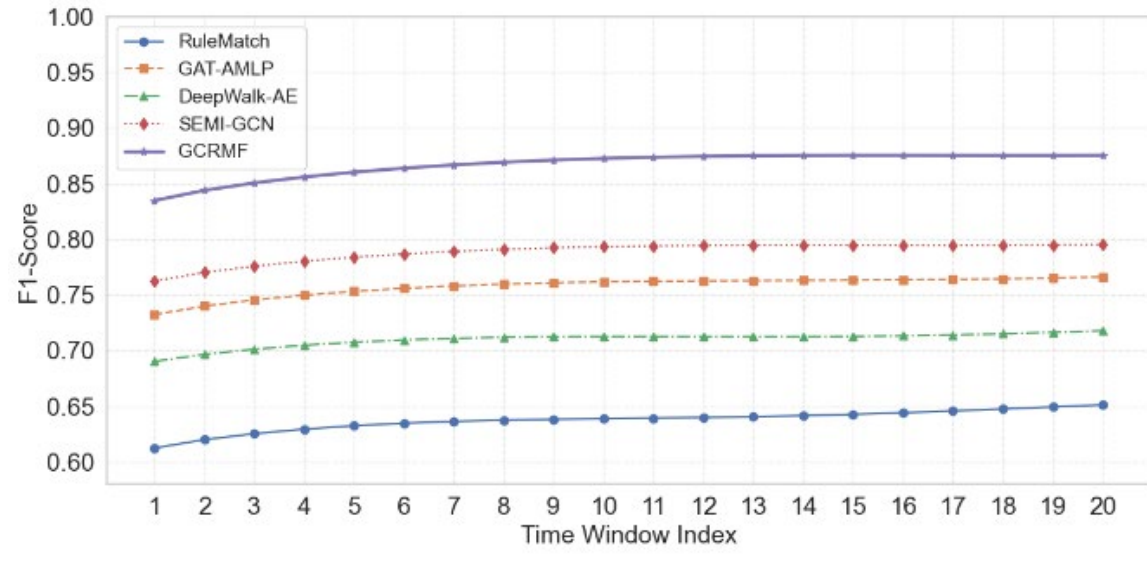

**Figure 4:** **F1-Score Comparison Across Methods**

Precision@K reflects the model's 'high-risk responsiveness' in actual warning scenarios, that is, how many of the top K suspicious transactions prioritized for review are indeed money laundering activities. It is suitable for risk control practices with limited resources.

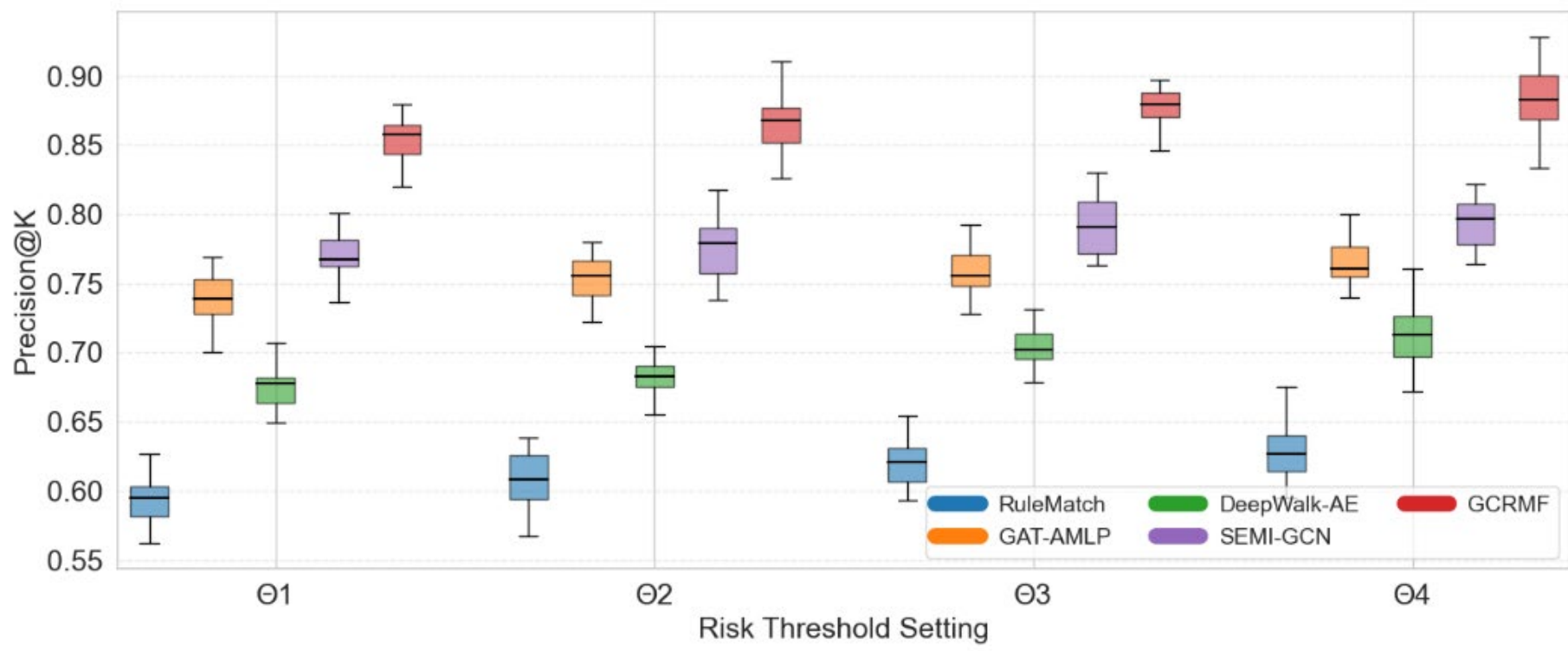

**Figure 5: Precision@K Comparison under Different Risk Thresholds**

As shown in Figure 5, under different risk threshold settings, the high-confidence warning accuracy of each method generally shows a trend of 'slightly increasing as the threshold rises.' Among them, the median and upper quartile of the traditional rule-based method RuleMatch are always the lowest, indicating that relying solely on manual rules makes it difficult to accurately screen the top K key inspection targets. DeepWalk-AE shows some improvement compared to RuleMatch, but limited by the expressive capability of unsupervised embeddings, its distribution remains relatively low and fluctuates considerably. GAT-AMLP and SEMI-GCN benefit from graph structure learning, with medians significantly higher than the first two, demonstrating stronger suspicious node ranking capabilities. In contrast, GCRMF achieves the highest median and upper quartile Precision@K across all thresholds, with a more concentrated box range. Niu et al. [19] systematically compared explainable AI algorithms, demonstrating their critical value for transparent and reliable decision-making in high-stakes domains.

# 5 Conclusion

In conclusion, we presented a framework named GCRMF that targets cross-industry mobility–energy supply chain networks, forms a Cross-Industry Heterogeneous Graph (CIHG), and utilizes dual-channel temporal graph attention encoding and meta-path subgraph reasoning together with contrastive self-supervised and online updating mechanisms to jointly model capital flow complexity and temporal evolution patterns to enhance detection of cross-industry recurring money laundering schemes. We are excited to enrich with real mobility–energy business data and multimodal business information, integrate deeper with large-scale graph pretraining models, and to explore interpretable subgraph discovery and federated learning mechanism and evaluate online deployment in regulatory sandbox as well as real financial institution environments. Adaptability under varying regional regulatory constraints requires further exploration, as compliance policies differ across jurisdictions. In future work, we aim to integrate additional data modalities, such as social media sentiment analysis, IoT-based energy usage patterns, and mobility behavior signals, to enhance the granularity of risk detection and enrich contextual understanding of suspicious transactions.